\documentclass{article}
\usepackage[preprint]{spconf}
\usepackage{amsmath,epsfig}
\usepackage{hyperref}
\usepackage{graphics}
\usepackage{graphicx}
\usepackage{amsfonts} 
\usepackage{amsmath}
\usepackage{amssymb}
\usepackage{bbm}
\usepackage{float}
\usepackage{xcolor}
\usepackage{booktabs}
\usepackage{multirow}
\usepackage{multicol}
\usepackage{microtype}
\usepackage{subcaption}
\usepackage{xspace}
\usepackage{tabularx}
\newcolumntype{Y}{>{\centering\arraybackslash}X}
\usepackage[capitalize]{cleveref}
\usepackage[numbers]{natbib}
\usepackage{comment}

\let\OLDthebibliography\thebibliography
\renewcommand\thebibliography[1]{
  \OLDthebibliography{#1}
  \setlength{\parskip}{0pt}
  \setlength{\itemsep}{0pt plus 0.3ex}
}

\newcommand{\ul}[1]{\underline{#1}} 
\newcommand{\eg}{\emph{e.g.}\xspace}
\newcommand{\ie}{\emph{i.e.}\xspace}

\begin{document}\sloppy

\def\x{{\mathbf x}}
\def\L{{\cal L}}

\title{Wasserstein loss for semantic editing in the latent space of GANs}
%
\name{Perla Doubinsky \hspace{0.3em} Nicolas Audebert \hspace{0.3em} Michel Crucianu \hspace{0.3em} Hervé Le Borgne}
\address{Preprint}


\maketitle

\begin{abstract}
The latent space of GANs contains rich semantics reflecting the training data. Different methods propose to learn edits in latent space corresponding to semantic attributes, thus allowing to modify generated images. Most supervised methods rely on the guidance of classifiers to produce such edits. However, classifiers can lead to out-of-distribution regions and be fooled by adversarial samples.
We propose an alternative formulation based on the Wasserstein loss that avoids such problems, while maintaining performance on-par with classifier-based approaches. We demonstrate the effectiveness of our method on two datasets (digits and faces) using StyleGAN2.
\end{abstract}

\section{Introduction}
\label{sec:intro}
GANs are known to encode the semantics of the training data in their latent space \cite{shen2020interfacegan,harkonen2020ganspace,shen2021closedform}.
Moving the latent codes in certain directions  results in changing specific semantic attributes 
in the generated images \cite{shen2020interfacegan}. This ability makes GANs great tools to perform image editing, especially as it can be applied to real images through inversion methods \cite{richardson2021encoding}.

\begin{figure*}[!t]
\centering
\begin{subfigure}{0.51\linewidth}
\begin{minipage}{0.5\linewidth}
\setlength\tabcolsep{2pt}
\begin{tabularx}{\linewidth}{cYYY}
 \hspace{1pt}  &  
 \small confidence ``Female'':
 & \vspace{4pt} 100\% & \vspace{4pt} 100\%
\end{tabularx}
\raisebox{0.2in}{\rotatebox[origin=c]{90}{\small LT \cite{yao2021latent}}} \includegraphics[width=4cm]{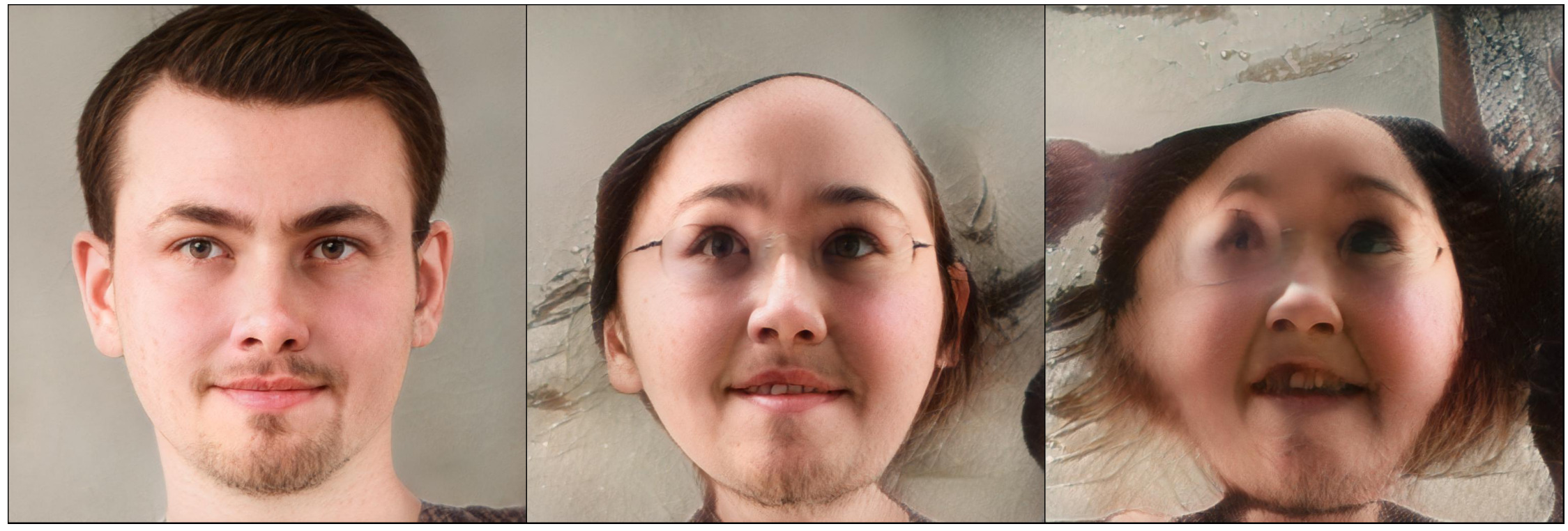} \\
\raisebox{0.2in}{\rotatebox[origin=c]{90}{\small Ours}} \hspace{1pt}\includegraphics[width=4cm]{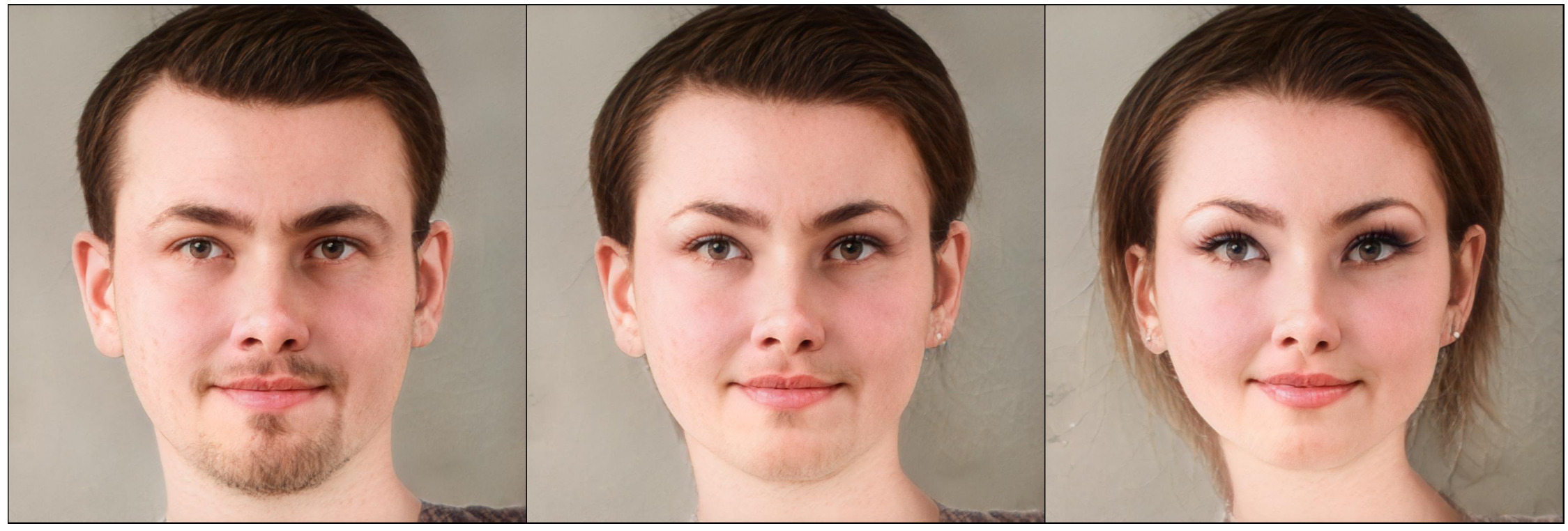} \\
\begin{tabularx}{\linewidth}{cYYY}
    \hspace{1pt} & Input & $\alpha=1$ & $\alpha=2$
\end{tabularx}
\end{minipage}
\begin{minipage}{0.45\linewidth}
\vspace{2em}
\includegraphics[width=3.5cm]{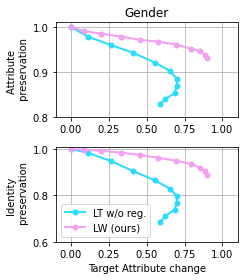}\\
\end{minipage}
\caption{FFHQ \cite{Karras2019stylegan2}, 'Male' $\rightarrow$ 'Female'}
\label{fig:limitation_lc_a}
\end{subfigure}
\begin{subfigure}{0.45\linewidth}
\begin{minipage}{0.38\linewidth}
\begin{tabularx}{\linewidth}{Yc}
\small confidence ``2 digits'':  & \raisebox{-12pt}{97 \%} \\
\end{tabularx}
\raisebox{0.2in}{\rotatebox[origin=c]{90}{\small LT \cite{yao2021latent}}} \includegraphics[height=1.4cm]{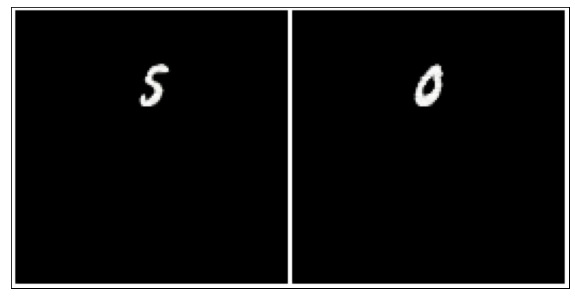}
\raisebox{0.2in}{\rotatebox[origin=c]{90}{\small Ours}} \includegraphics[height=1.4cm]{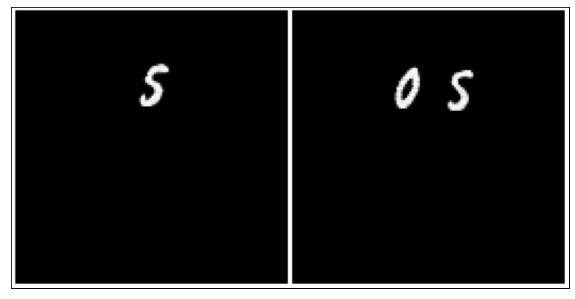} 
\begin{tabularx}{\linewidth}{cYY}
    & Input & $\alpha=1$ 
\end{tabularx}\vspace{12pt}
\end{minipage}
\begin{minipage}{0.4\linewidth}
\includegraphics[width=3.8cm]{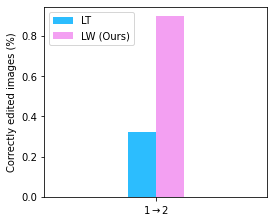}
\end{minipage}
\caption{MultiMNIST \cite{mulitdigitmnist}, '1 digit' $\rightarrow$ '2 digits'}
\label{fig:limitation_lc_b}
\end{subfigure}
\caption{
\textbf{Failure cases of a classifier-based method.} LT~\cite{yao2021latent} learns edits in latent space under the guidance of a latent classifier. (a) On FFHQ: without $L2$-regularization on the edited codes, the edited images are unrealistic (as shown in the qualitative result on the left) before reaching the desired editing. The classifier leads to out-of-distribution regions as it allocates high confidence to regions larger than that of the training samples \cite{nguyen2015deep}. The quantitative analysis on attribute and identity preservation shows highly degraded results. (b) On MultiMNIST: the edited images remain unchanged (no digit is being added) while the classifier indicates the opposite (predicts 2 digits with high confidence).
The classifier leads to regions close to the decision boundaries where there are adversarial samples.
The quantitative analysis shows that only 32\% of images are correctly edited.
}
\label{fig:limitation_lc}
\end{figure*}

The challenge is to identify the manipulations in the latent space that have the desired effect on one attribute without affecting others. To obtain such \emph{disentangled} manipulations, existing supervised methods leverage the semantic knowledge learned by pretrained attribute classifiers operating either in the image domain (\emph{image} classifiers) or directly in the latent domain (\emph{latent} classifiers). The key idea is that manipulated latent codes (or the images they produce) shift the predictions to match the desired outcome \cite{hou2022guidedstyle,yao2021latent}.
However, classifiers can easily be fooled \cite{nguyen2015deep}, \eg they can classify with high confidence out-of-distribution samples.
As illustrated in \cref{fig:limitation_lc_a}, the latent classifier of \cite{yao2021latent} steers latent codes outside the distribution resulting in edited images that are unrealistic.
To address this issue, the authors 
employ an \emph{ad hoc} $L2$-regularization to minimize the norm of the latent editing.
While this fixes out-of-distribution edits, \cref{fig:limitation_lc_b} shows that on MultiMNIST \cite{mulitdigitmnist} this regularization produces adversarial samples \cite{szegedy2013intriguing} instead, \emph{i.e.}\ the edited latent codes are correctly classified but the corresponding images remain unchanged. This is not surprising as changing the predicted class while minimizing the $L2$-norm of the edit precisely mimics the search for adversarial examples.
To overcome these issues, we introduce a new formulation for learning semantic editing in the latent space, leading to a core solution that \emph{does not} rely on classifiers. 
\begin{figure}[t]
    \centering
    \includegraphics[width=8cm]{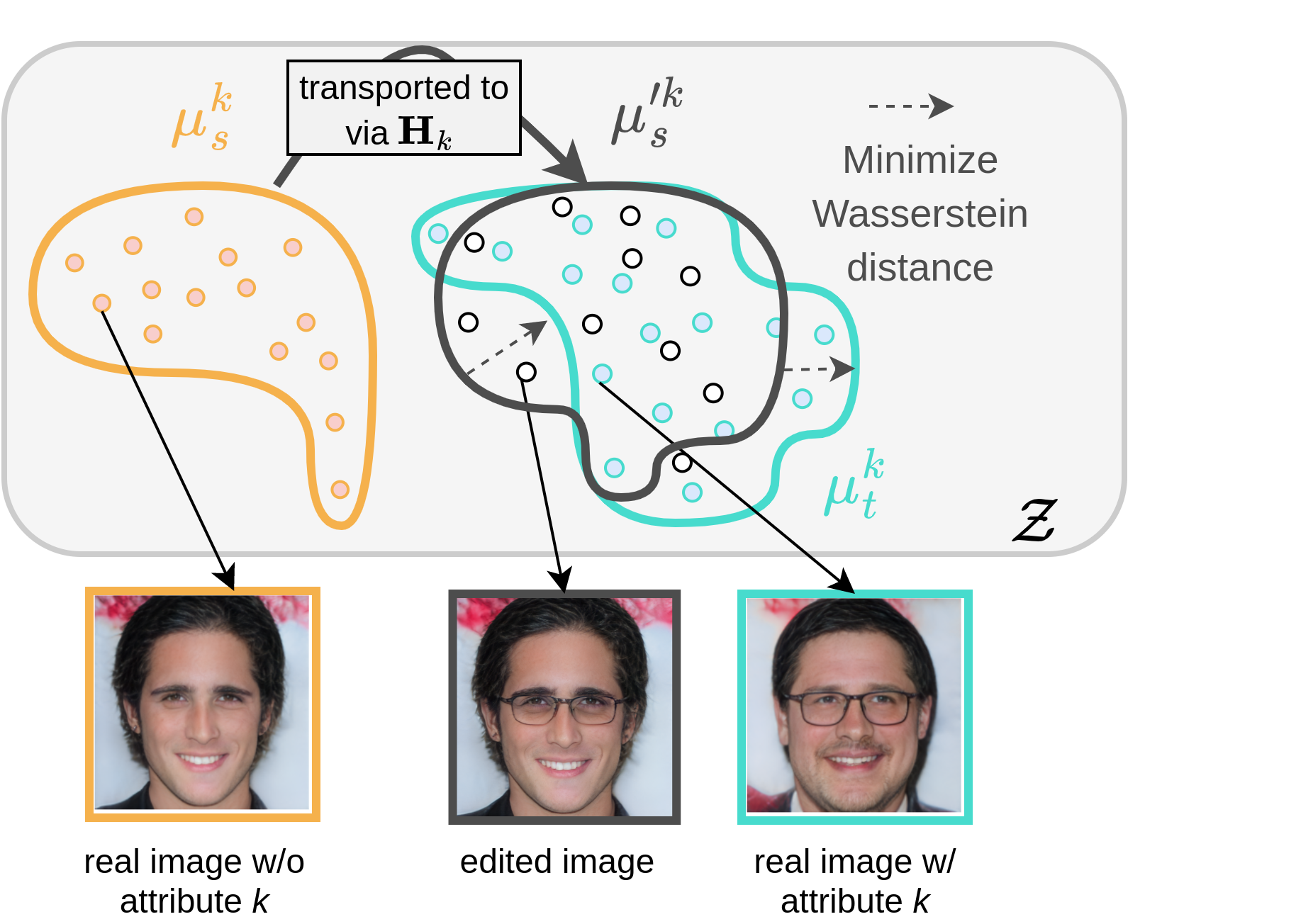}
    \caption{
    \textbf{Method overview.} 
    For each semantic attribute (\emph{e.g.}\ ``Glasses'') we learn a mapping $\mathbf{H}_k$ that moves the distribution of latent codes lacking the attribute to the distribution of codes having that attribute. We enforce that each latent code is moved near a point that shares similar semantics, thus only changing that attribute. 
    For identity preservation, the resulting distribution does not entirely match the target distribution. 
    }
    \label{fig:my_label}
\end{figure}

From a global perspective, latent editing can be viewed as an optimal transport problem \cite{Villani2008OptimalTO}.
Given a distribution of latent codes sharing some semantics, we propose to transport it onto the distribution of latent codes that share the same semantics except for the attribute to be edited. Since the resulting images should not exhibit any other changes than the desired one, the initial points should be transported ``close'' to points sharing their semantics; that is, the transport should be optimal w.r.t.\ a cost representing the perceptual similarity.
To achieve this, we learn transformations in latent space using the guidance of the Wasserstein loss with an Euclidean cost, 
which can be combined with a Wasserstein loss with a cost computed in the attribute space to enforce disentanglement. 

We apply our method in the latent space of StyleGAN2 to modify the number of digits and edit facial attributes. 
We compare quantitatively and qualitatively to the method of Yao~\textit{et al}. (LT) \cite{yao2021latent} that relies exclusively on a latent classifier. 
Without additional regularization, our method leads to realistic edited images and achieves on-par disentanglement and 
identity preservation than a classifier-based method. 

\section{Related work}
\label{sec:related_work}
Early works on GANs have demonstrated that their latent space contains rich semantics that can be leveraged 
to control some properties of the generated data. Simply translating a latent code in a given direction can lead to the variation of a semantic attribute in the corresponding generated image \cite{shen2020interfacegan,harkonen2020ganspace,shen2021closedform,spingarn2020gan}. Latent semantic directions can be 
extracted from the latent space without supervision by performing PCA 
\cite{harkonen2020ganspace} or by singular value decomposition on the weights of the pretrained GAN \cite{shen2021closedform, spingarn2020gan}. Supervised methods often employ classifiers to extract the directions. InterfaceGAN \cite{shen2020interfacegan} introduces a framework to edit binary facial attributes. An SVM is trained in latent space to infer the hyperplane that best separates the positive vs.\ negative latent codes w.r.t.\ a semantic attribute. The vector orthogonal to the hyperplane then constitutes the editing direction. Later works aim at learning a direction specific to each latent code by passing the input code through an MLP or an affine layer that is trained with the guidance of a classifier. 
GuidedStyle \cite{hou2022guidedstyle} uses an attribute image classifier that classifies the images corresponding to the edited latent codes. The editing is correct if the classifier's predictions correspond to the desired change. Yao et al.~\cite{yao2021latent}
employ a similar objective but use a classifier trained directly in latent space. 
However, classifiers are unreliable \cite{szegedy2013intriguing,nguyen2015deep}, potentially leading to images or latent codes that minimize the objective but do not correspond to the desired editing. 
Different from previous works, our core method does not rely on classifiers. Instead, we solve the problem using the optimal transport framework. 
To the best of our knowledge, this is the first work applying optimal transport for latent space editing.

\section{Wasserstein loss for GAN editing} 

Let $G$ be a pretrained generator and $\mathcal{Z}$ its latent space such that $\mathbf{I} = G(\mathbf{z})$ where $\mathbf{z} \in \mathcal{Z}$ is a latent code and $\mathbf{I}$ the corresponding generated image. Suppose we have a collection of latent codes $\{\mathbf{z}^{(i)}\}_{i=1}^N$, where each code is associated with a set of binary semantic attributes $\mathcal{A}=\{\mathbf{a}_1$, $\mathbf{a}_2$, .., $\mathbf{a}_K\} \in \{0,1\}$. For a given attribute $\mathbf{a}_k$, we aim to learn an affine transform $\mathbf{H}_k$ in $\mathcal{Z}$,
\begin{equation}
    \mathbf{z}'_k = \mathbf{z}_k + \alpha \cdot \mathbf{H}_k(\mathbf{z}), \text{ $\alpha \in \mathbb{R}$
    }
    \label{eq:model}
\end{equation}
such that only the attribute intensity $\mathbf{a}_k$ differs in the resulting image $\mathbf{I}' = G(\mathbf{z}')$,
where $\alpha$ controls the strength of the change.

Let $\mu_s^{k}$ be the distribution of latent codes $\mathbf{z}_k$ that are negative with respect to the binary attribute $\mathbf{a}_k$ and $\mu_t^{k}$ the distribution of latent codes $\mathbf{\bar{z}}_k$ positive w.r.t.\ the attribute $\mathbf{a}_k$.
To increase the intensity of the attribute $\mathbf{a}_k$ in the generated images, $\mathbf{H}_k$ should transport the distribution of edited latent codes $\mathbf{z}'_k$ denoted by $\mu'^k_s$ close to the distribution $\mu^k_t$. 
However, the information encoding other attributes or properties should remain unchanged.
The theory of optimal transport \cite{Villani2008OptimalTO} 
introduces a framework to transport a distribution to another with a minimal cost. The Wasserstein distance between two distributions represents the minimal value of this cost. Thus, we propose to use this loss as supervision to learn $\mathbf{H}_k$ with a cost in latent space expressing similarity in image space. 
We call this model Latent Wasserstein (LW).

\subsection{Wasserstein Distance}

Let us define two discrete distributions:
 \begin{equation}
 \mu_s = \sum_{i=1}^{n_s} a_i \delta({x_i}) \text{ and } \mu_t = \sum_{i=1}^{n_t} b_i \delta({y_i})
\label{eq:discrete_distrib}
\end{equation}
where $\delta(,)$ is the Dirac function and $a_i, b_i$ the probability mass associated with each sample.

\noindent The Wasserstein distance 
between $\mu_s$ and $\mu_t$ is defined as:
\begin{align}
    W(\mu_s, \mu_t) = \min \sum_{i,j} T_{i,j}c_{i,j} \nonumber \\ 
    \text{s.t. } T \mathbf{1}_{n_t} = \mu_s,  T^{\top} \mathbf{1}_{n_s} = \mu_t 
\end{align}
$T$ is the transport matrix. 
$T_{i,j}$ represents how much probability mass must be transported from point $x_i$ to point $y_i$ and $c_{i,j}$ the cost of this transport. Estimating the Wasserstein distance is challenging in practice as it requires to solve the underlying optimal transport. The Wasserstein distance is usually estimated with the Sinkhorn divergence built on entropic regularization with debiasing terms \cite{feydy2019interpolating,NIPS2013_af21d0c9}.

\subsection{Core Method}
\label{subsec:main_objective}

Our main objective is to minimize the Wasserstein loss between $\mu'^k_s$ and $\mu^k_t$ with a squared Euclidean cost function:
\begin{align}
   \mathcal{L}_\text{edit} = W\left(\mu'^k_s, \mu_t^k\right), \quad x_i = \mathbf{z'}_k^{(i)} \text{ and } y_j = \mathbf{\bar{z}}_k^{(j)}
   \nonumber \\
    c_{i,j} = \frac{1}{2}\rVert x_i - y_j\lVert^2 
    \label{eq:l_edit}
\end{align}

In \cref{eq:discrete_distrib}, the probability mass of each sample is usually set uniformly across samples, \ie $a_i=\frac{1}{n_s}$ and $b_i=\frac{1}{n_t}$ for all $i$. If there are biases in the collection of training latent codes, the representation of semantically similar samples may vary significantly between the $\mu'^k_s$ and $\mu^k_t$ \cite{doubinskybalancing2022}. 
In this case, we propose to weight the source samples according to the number of similar samples in the target distribution. More formally, we set  $a_i =\frac{1}{n^A_t \times n^A_s}$ where $n^A$ is the number of latent codes with the attribute combination $A$ for a set of selected attributes.


\subsection{Enforced Disentanglement}
\label{subsec:attr_disentanglement}
To ensure that the transported latent codes share the same attributes as the initial ones, we propose to minimize the Wasserstein loss between $\mu'^k_s$ and $\mu^k_s$: 

\begin{align}
   \mathcal{L}_\text{edit} = W\left(\mu'^k_s, \mu^k_s\right), \quad x_i = \mathbf{z'}_k^{(i)} \text{ and } y_j = \mathbf{z}_k^{(j)} \nonumber \\
    c_{i,j} = \frac{1}{2}\sum_{l \neq k} (1-\gamma_{lk})\rVert \mathbf{F}_l(x_i) -\mathbf{F}_l(y_j)\lVert_2^2
\label{eq:attribute_cost}
\end{align}


In contrast to the previous objective, we employ a cost computed in the \emph{attribute} space. We follow the cost defined in \cite{yao2021latent},  
where $\mathbf{F}_l$ is a latent classifier trained to predict $\mathbf{a}_l$ given a latent code $\mathbf{z}$. The term $\gamma_{lk}$ represents the absolute correlation between attribute $\mathbf{a}_l$ and $\mathbf{a}_k$ and is used to avoid disentangling naturally correlated attributes (\emph{e.g.}\ ``Smile'' and ``High Cheekbones''). 
Although we use the cost introduced in \cite{yao2021latent}, our constraint is a more relaxed constraint since we operate on the distribution. 

The final objective to minimize is then $\mathcal{L} = \mathcal{L}_\text{edit} + \lambda \mathcal{L}_\text{pres}$ where $\lambda$ allows to balance the two losses.

\section{Experiments}

\begin{figure*}[t]
    \centering
    \begin{tabular}{cccc}
    & Gender & Glasses & Blond Hair \\
             \includegraphics[width=2cm]{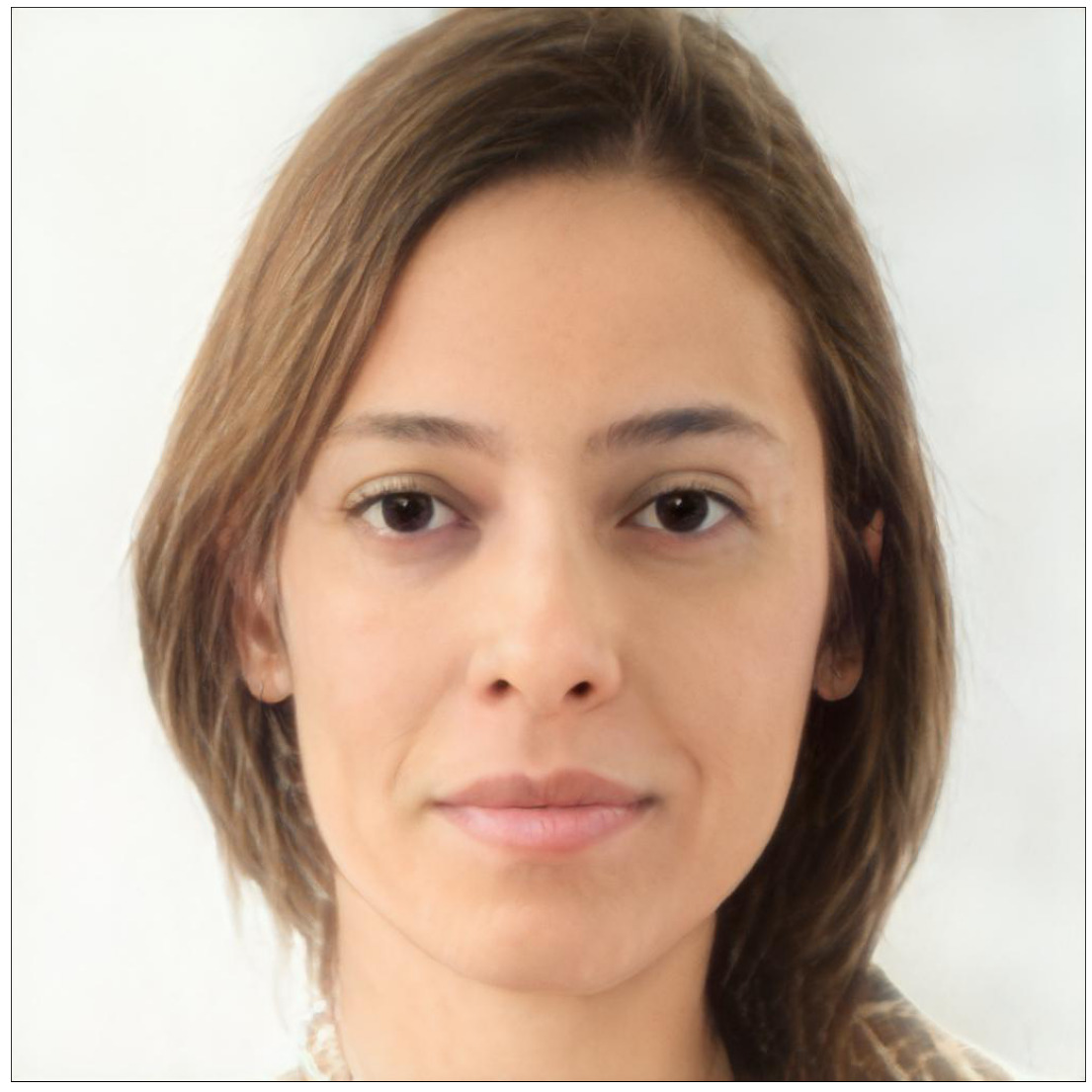} &
    \includegraphics[width=4cm]{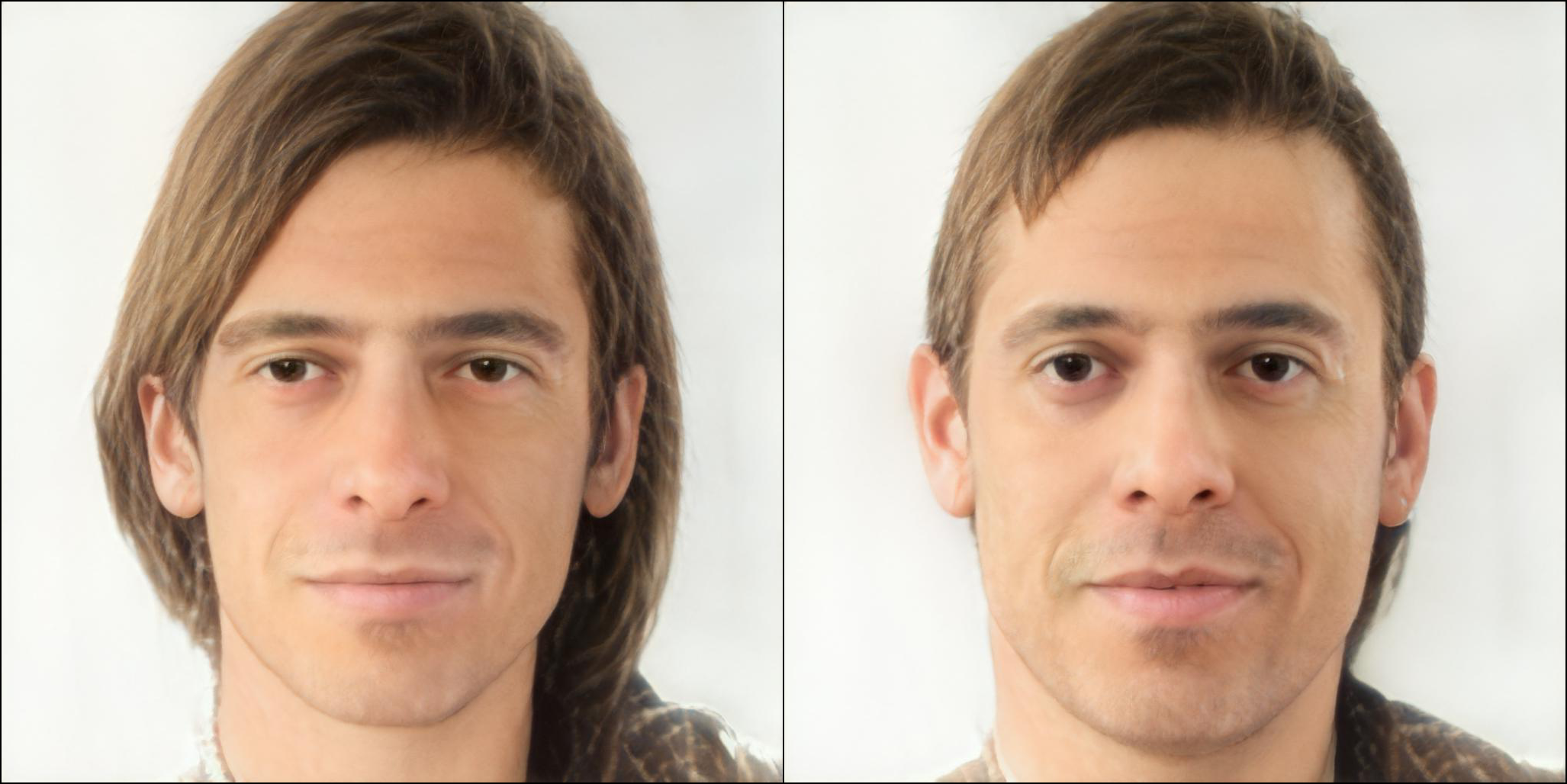} &
    \includegraphics[width=4cm]{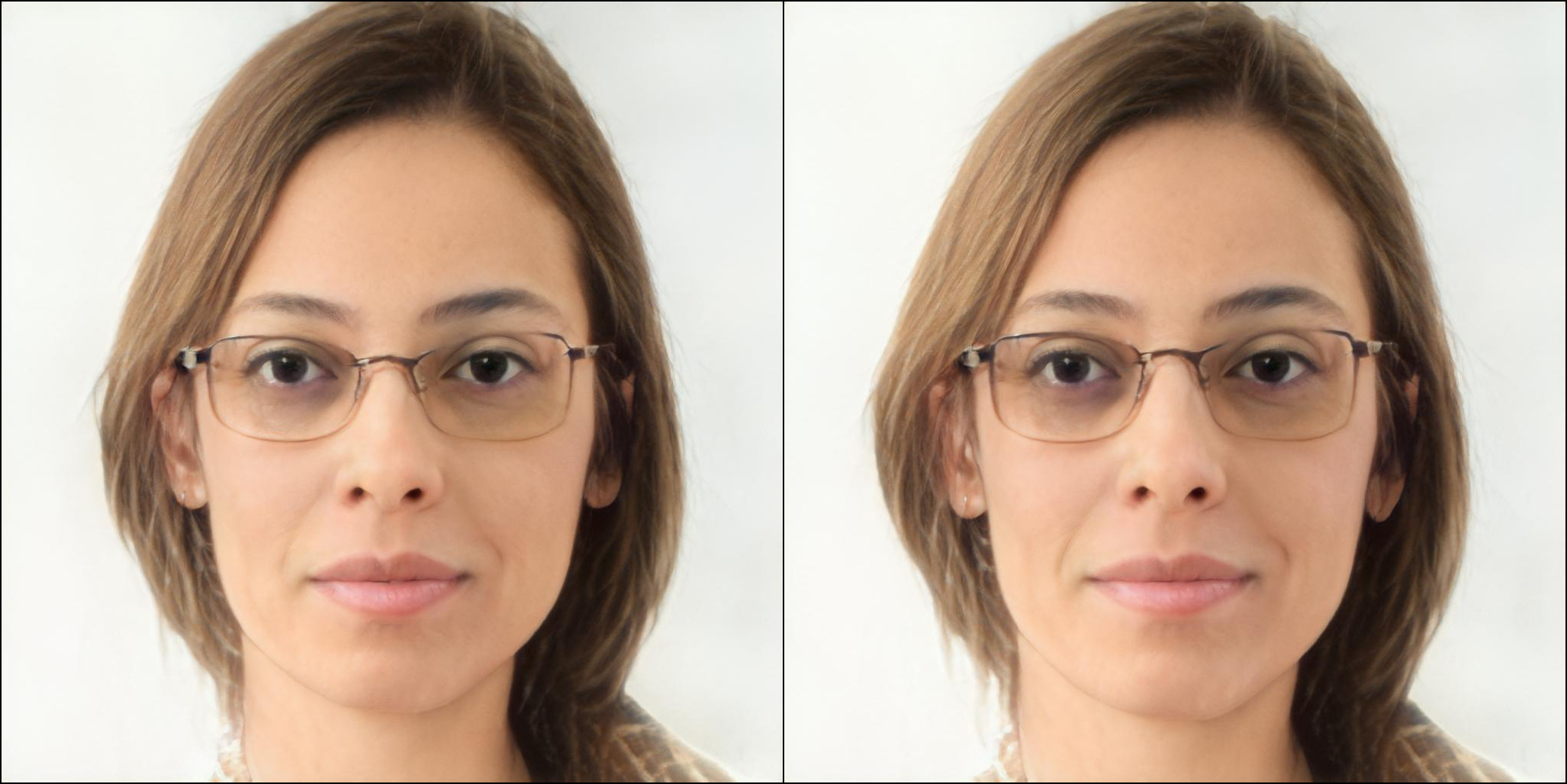} &
    \includegraphics[width=4cm]{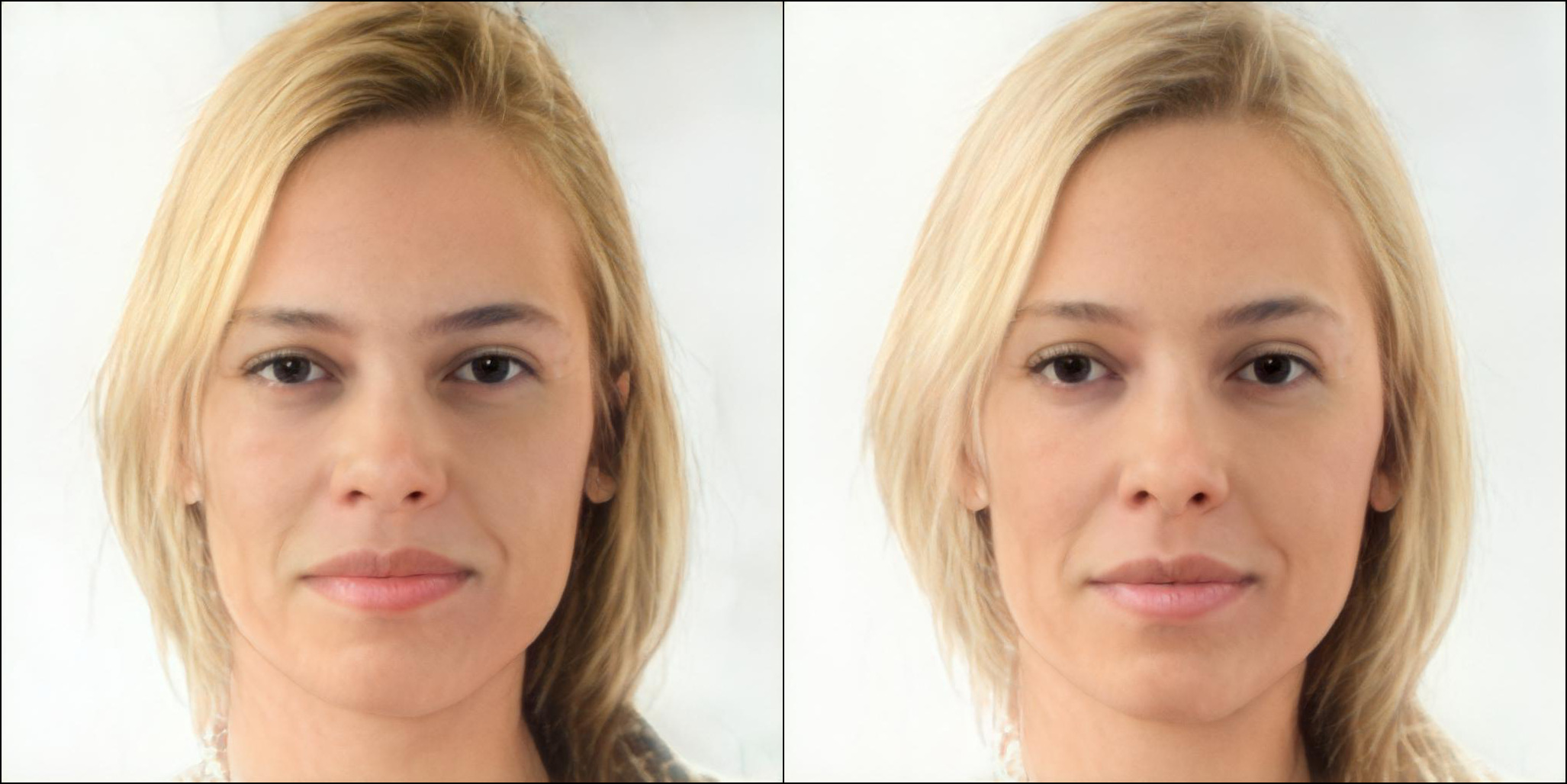} \\
         \includegraphics[width=2cm]{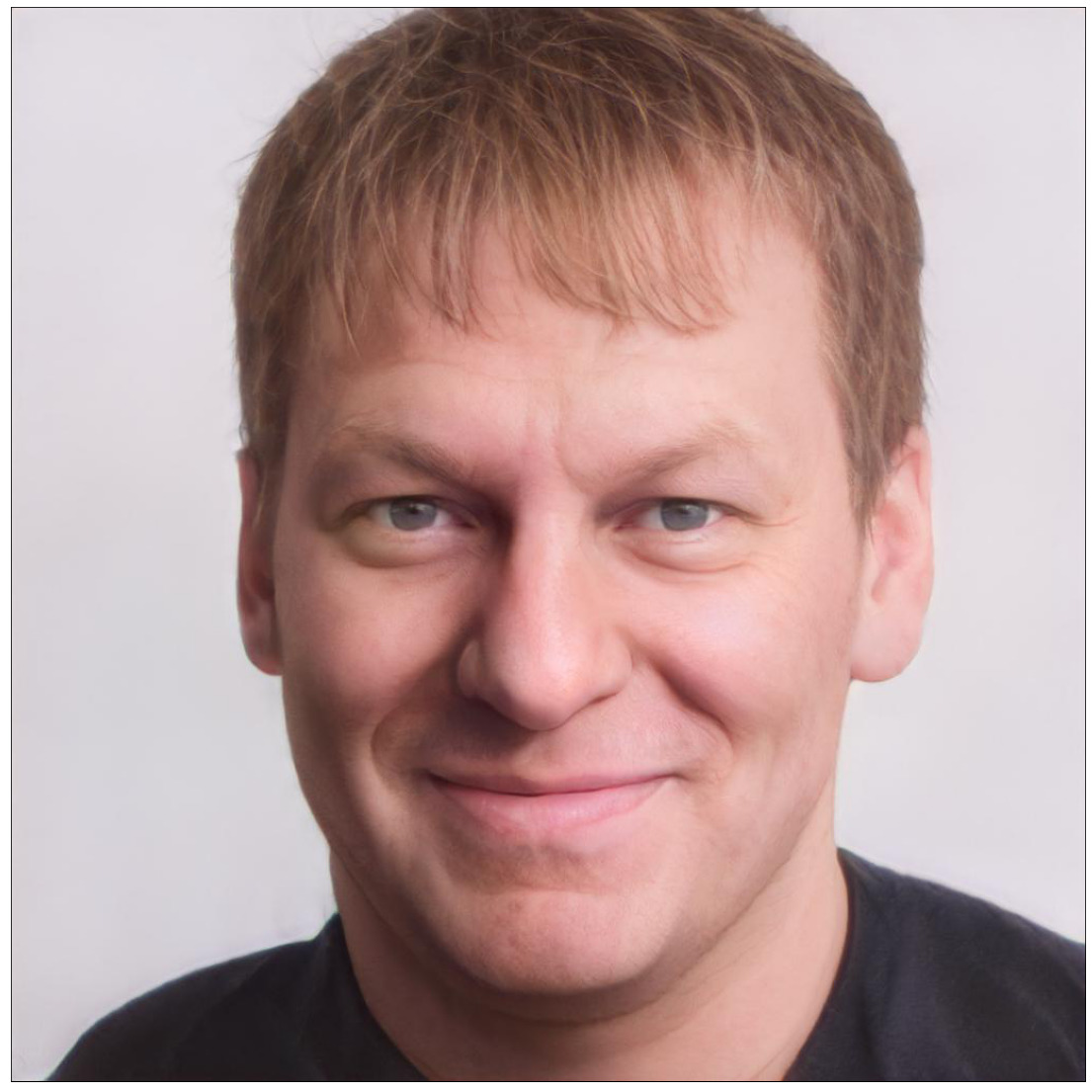}
    & \includegraphics[width=4cm]{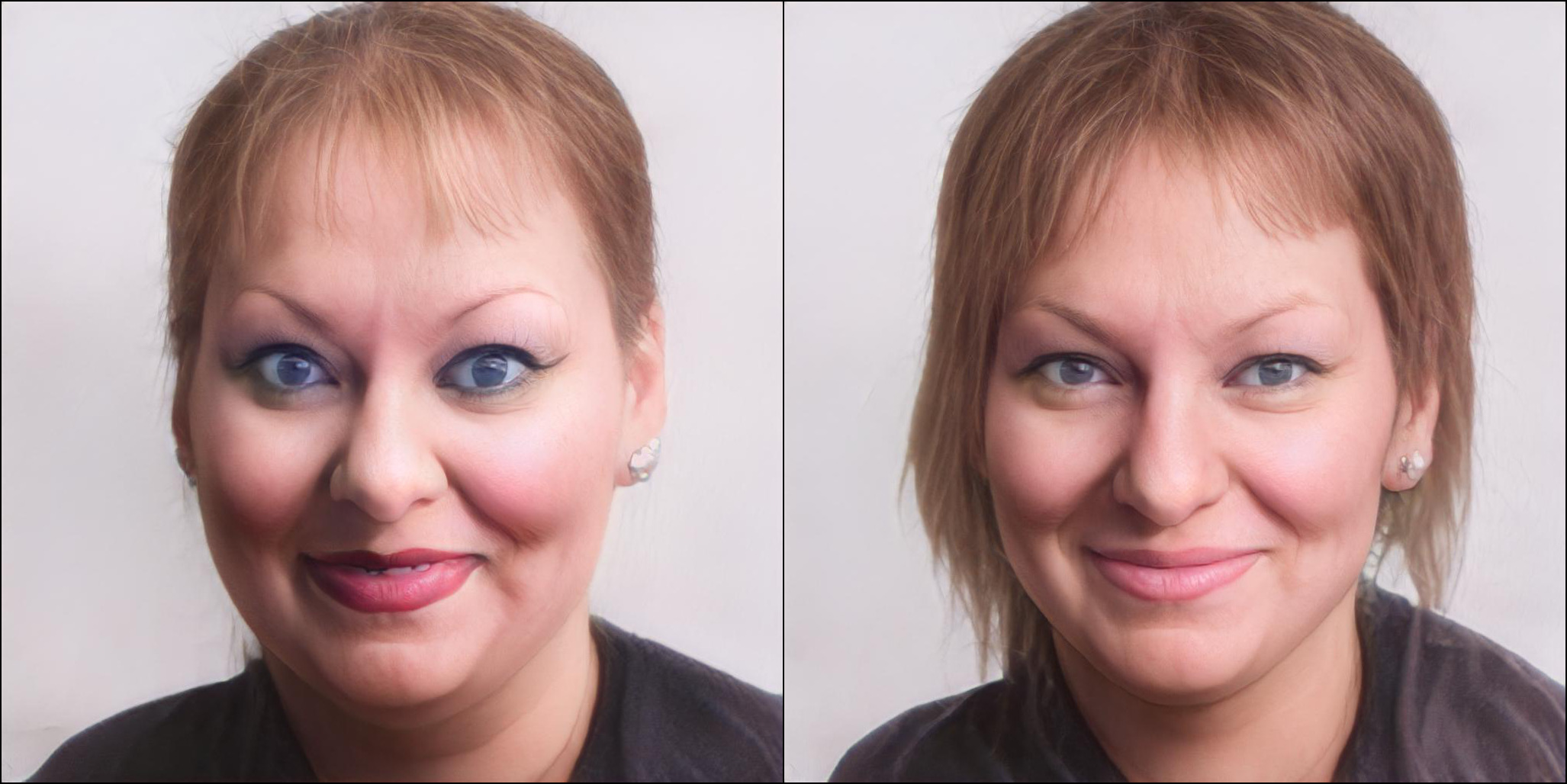} & 
    \includegraphics[width=4cm]{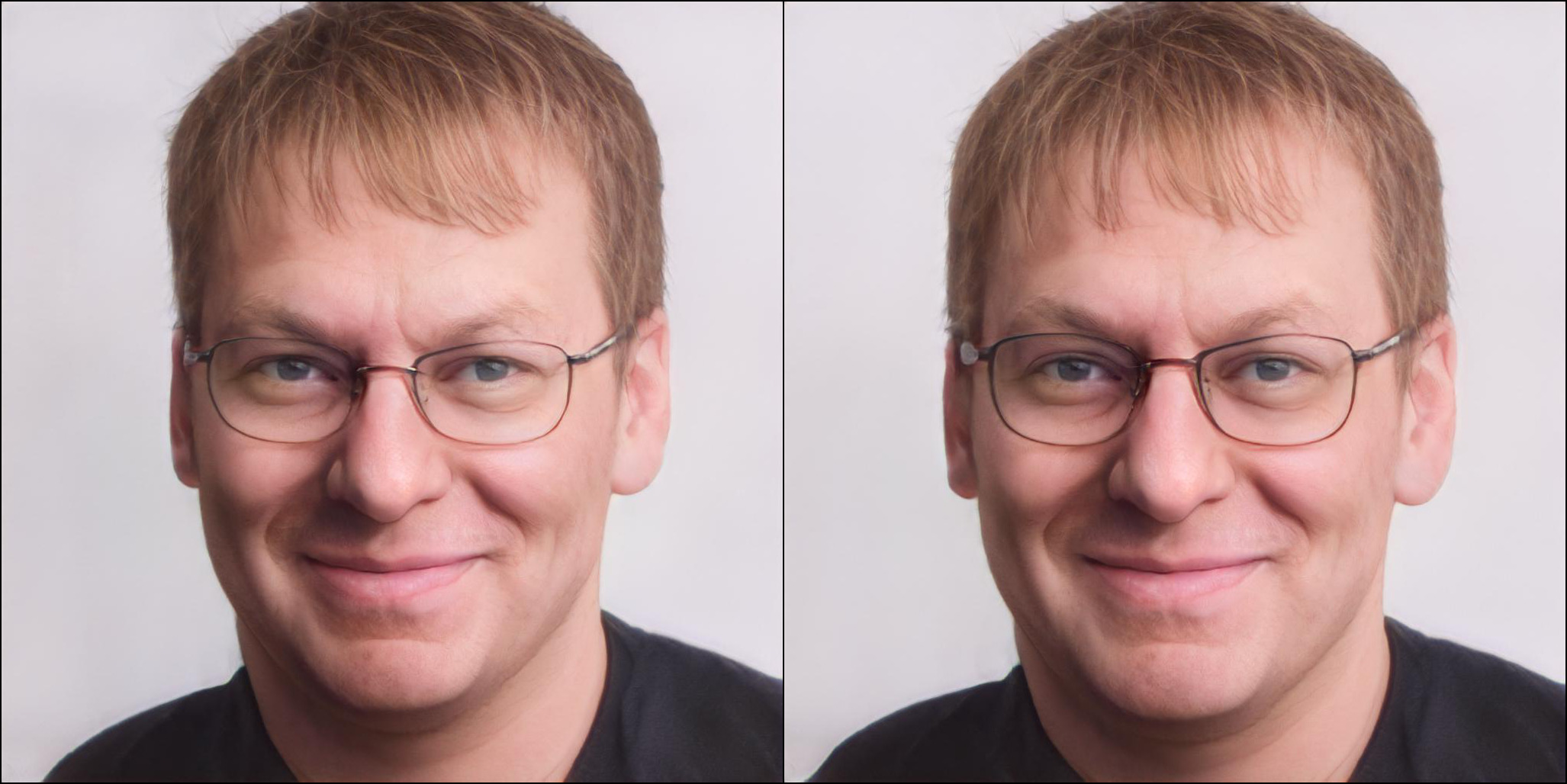} & 
    \includegraphics[width=4cm]{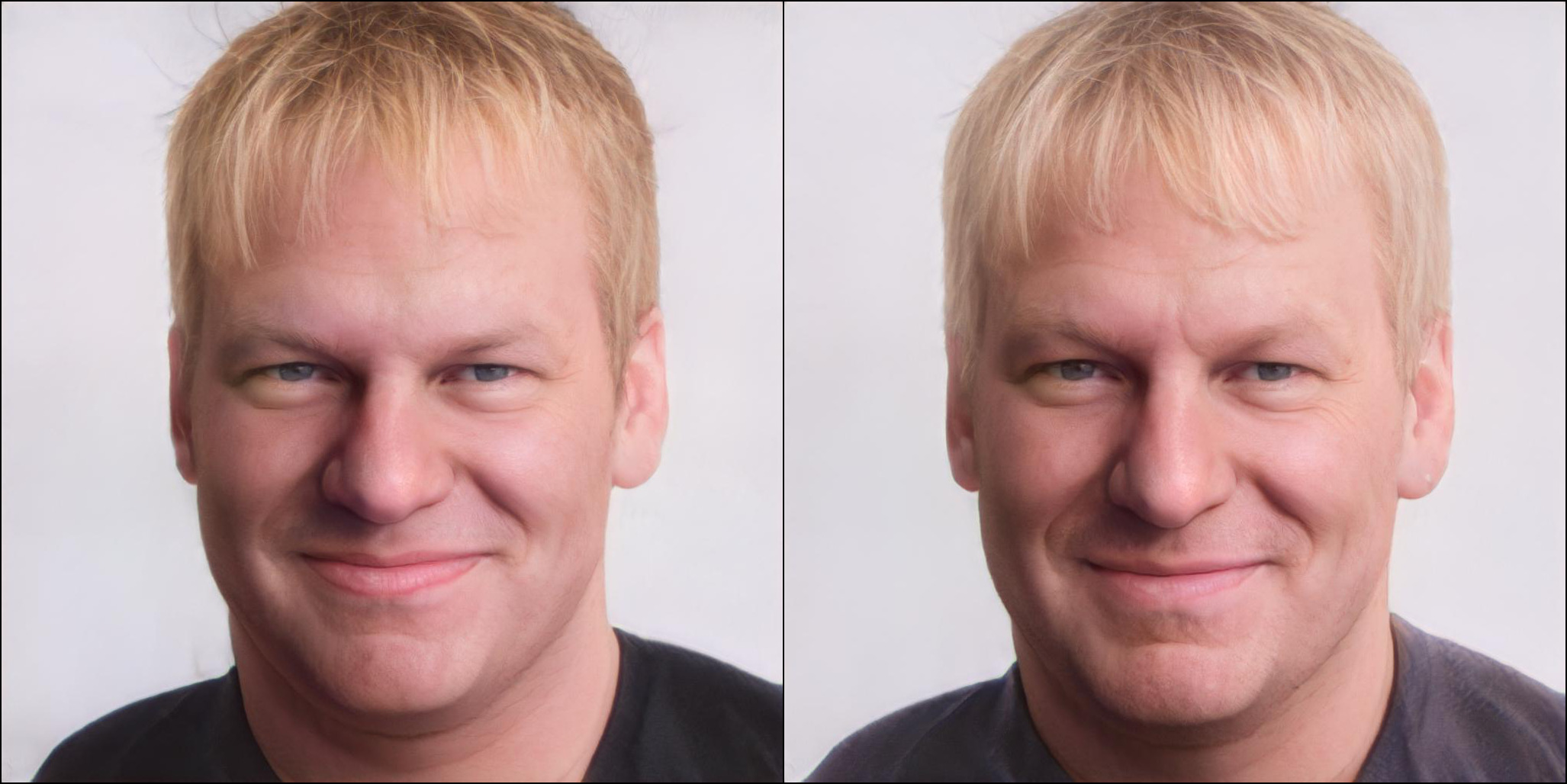} \\
    & Age & Bangs & Wavy Hair \\    
    \includegraphics[width=2cm]{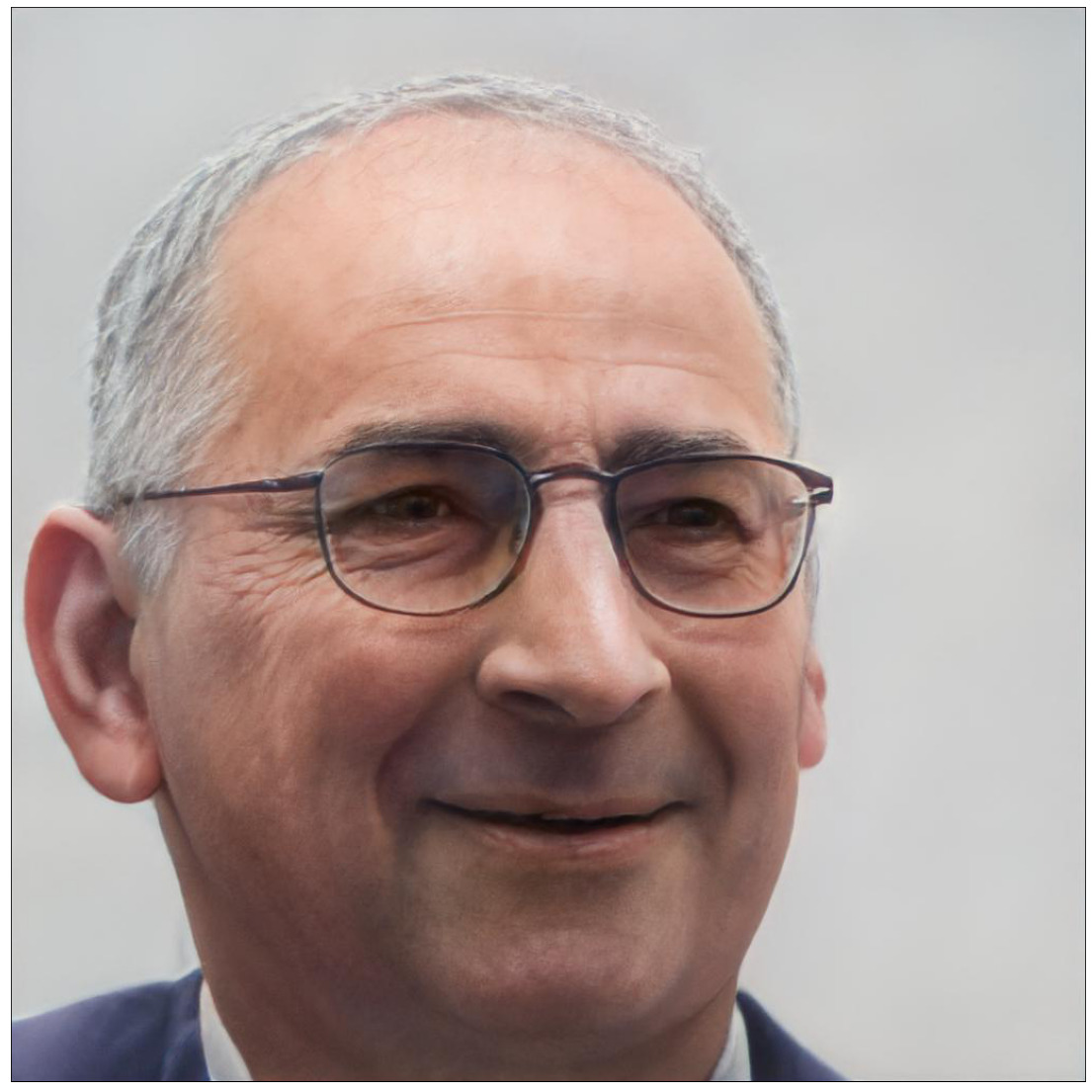} &    
    \includegraphics[width=4cm]{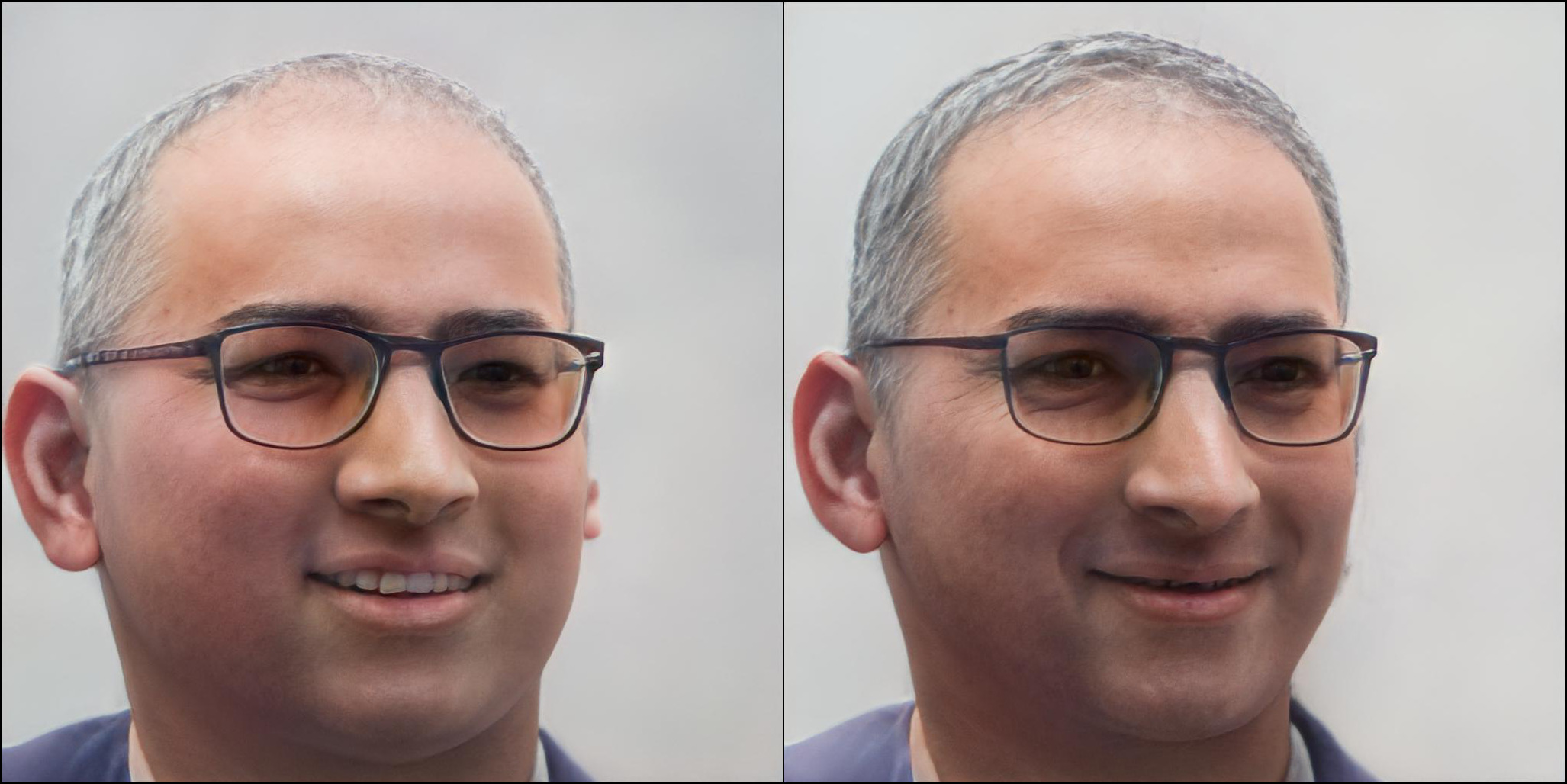} &
    \includegraphics[width=4cm]{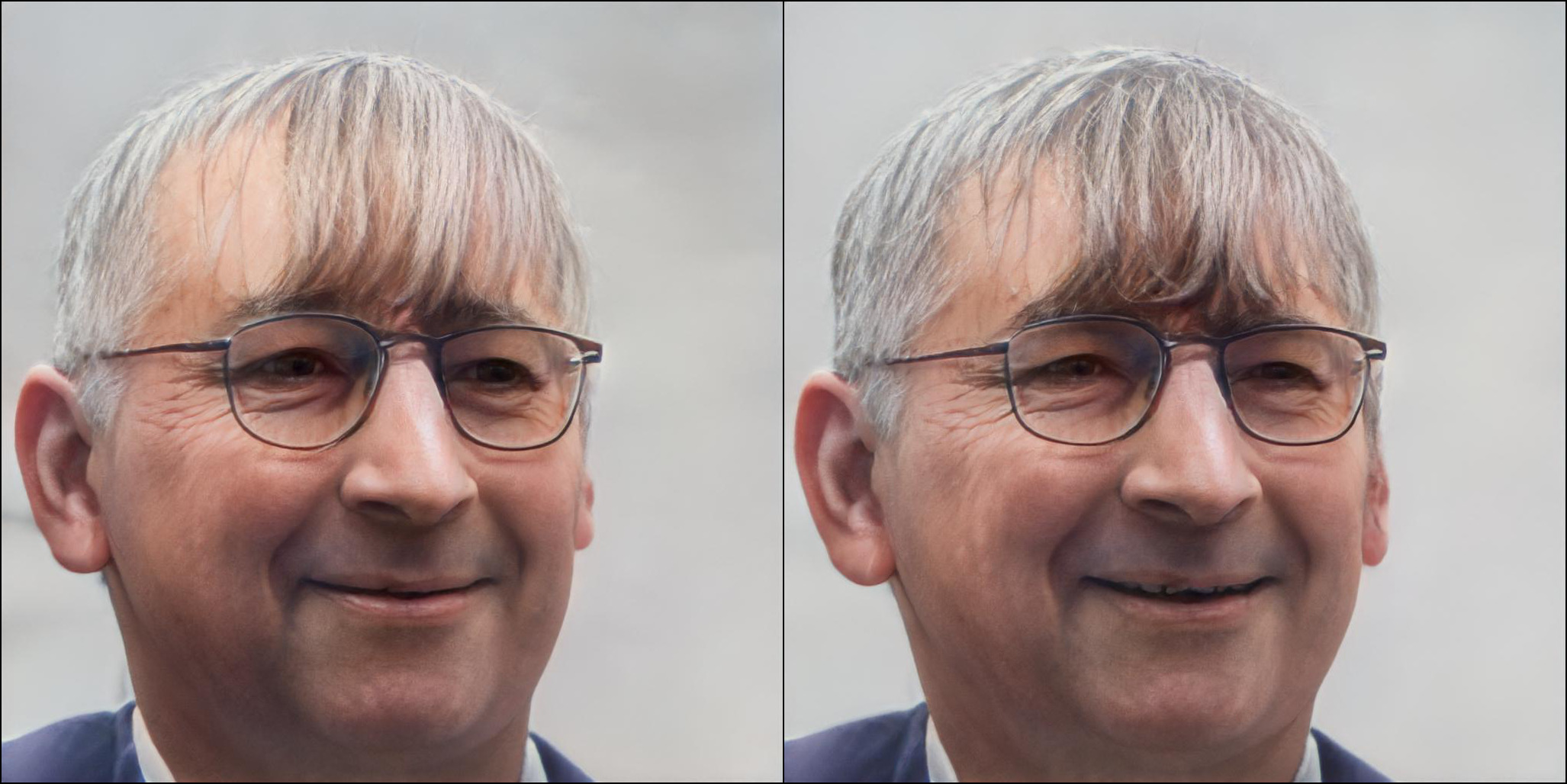} 
    & 
    \includegraphics[width=4cm]{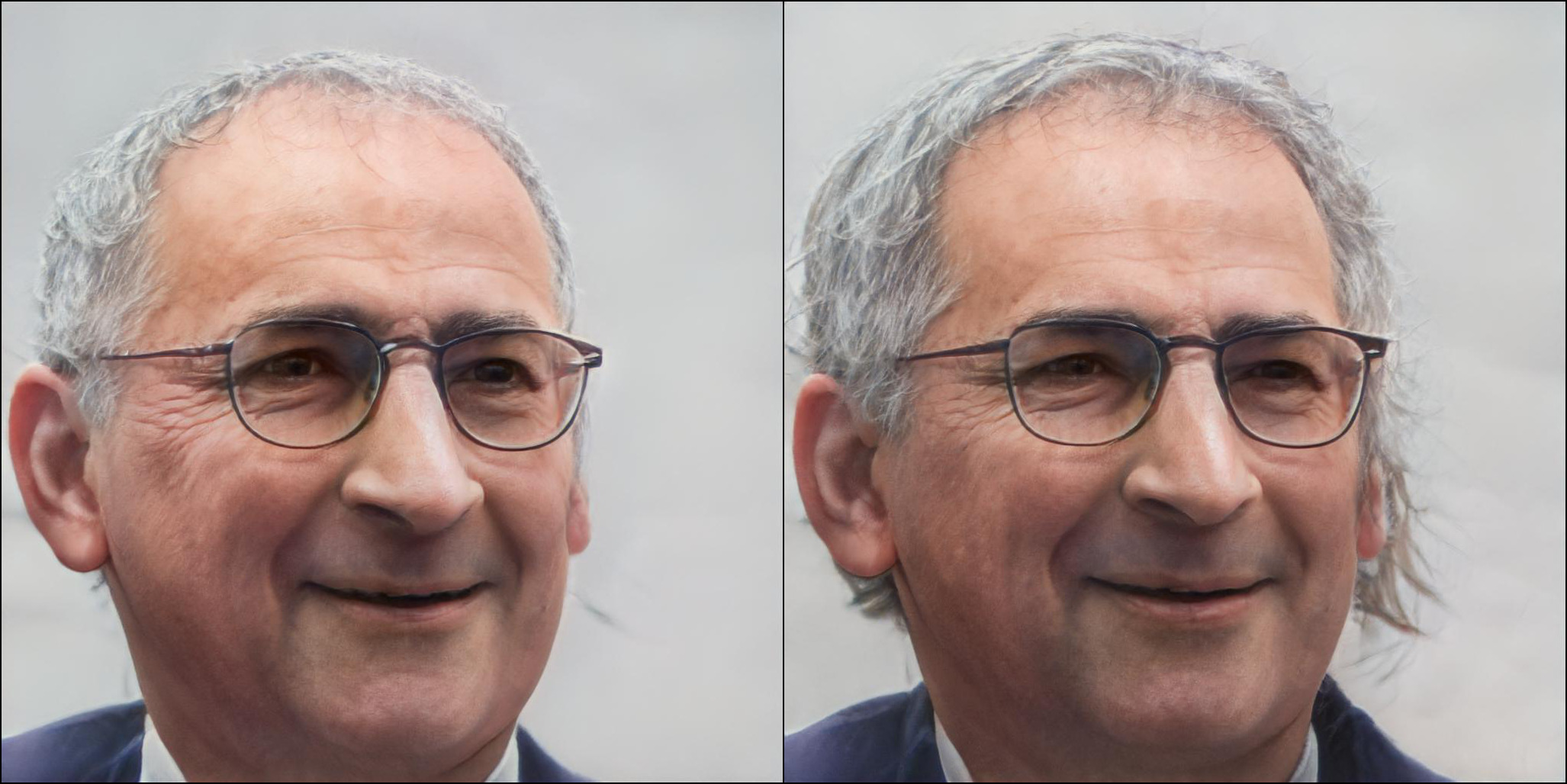} \\
    \includegraphics[width=2cm]{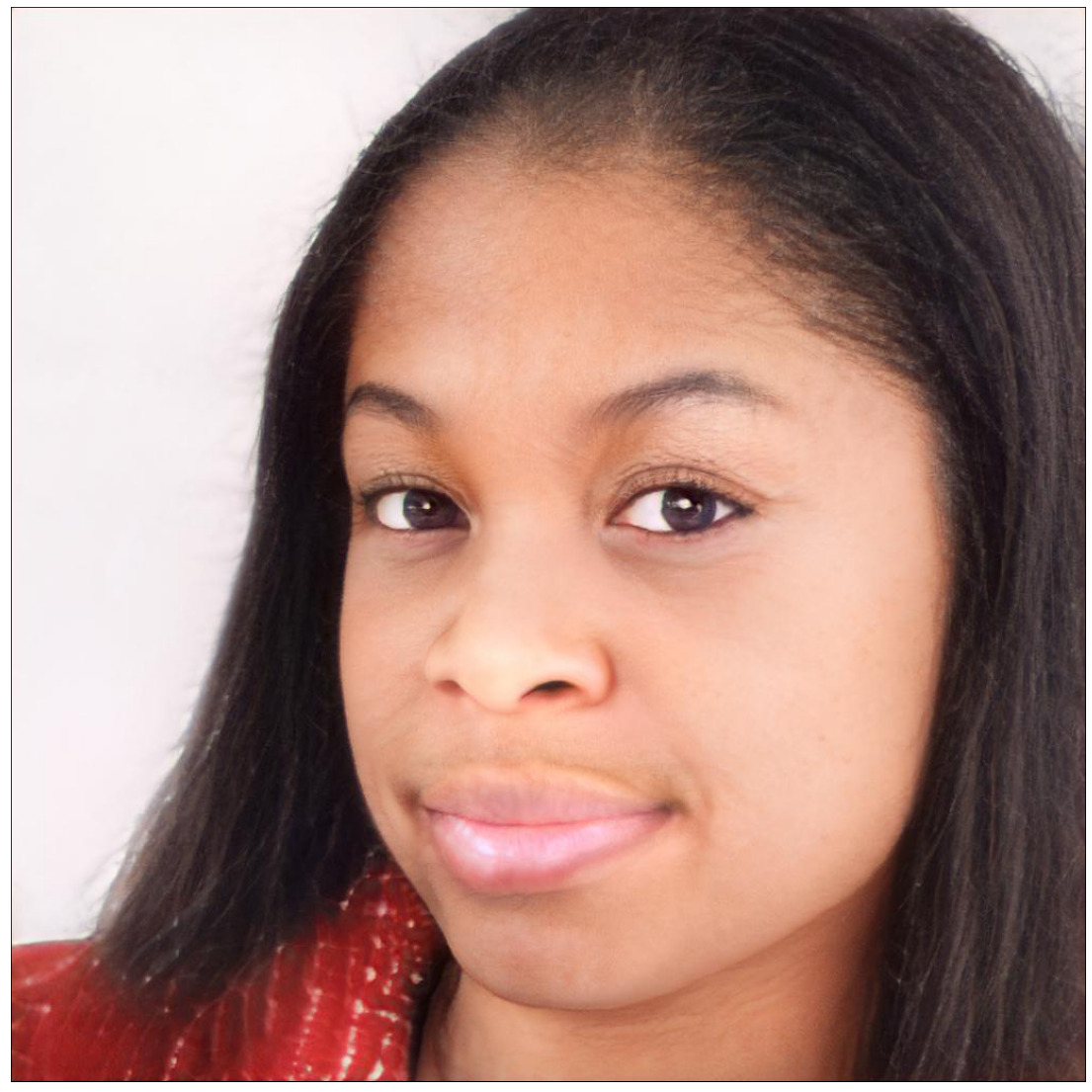} &    
    \includegraphics[width=4cm]{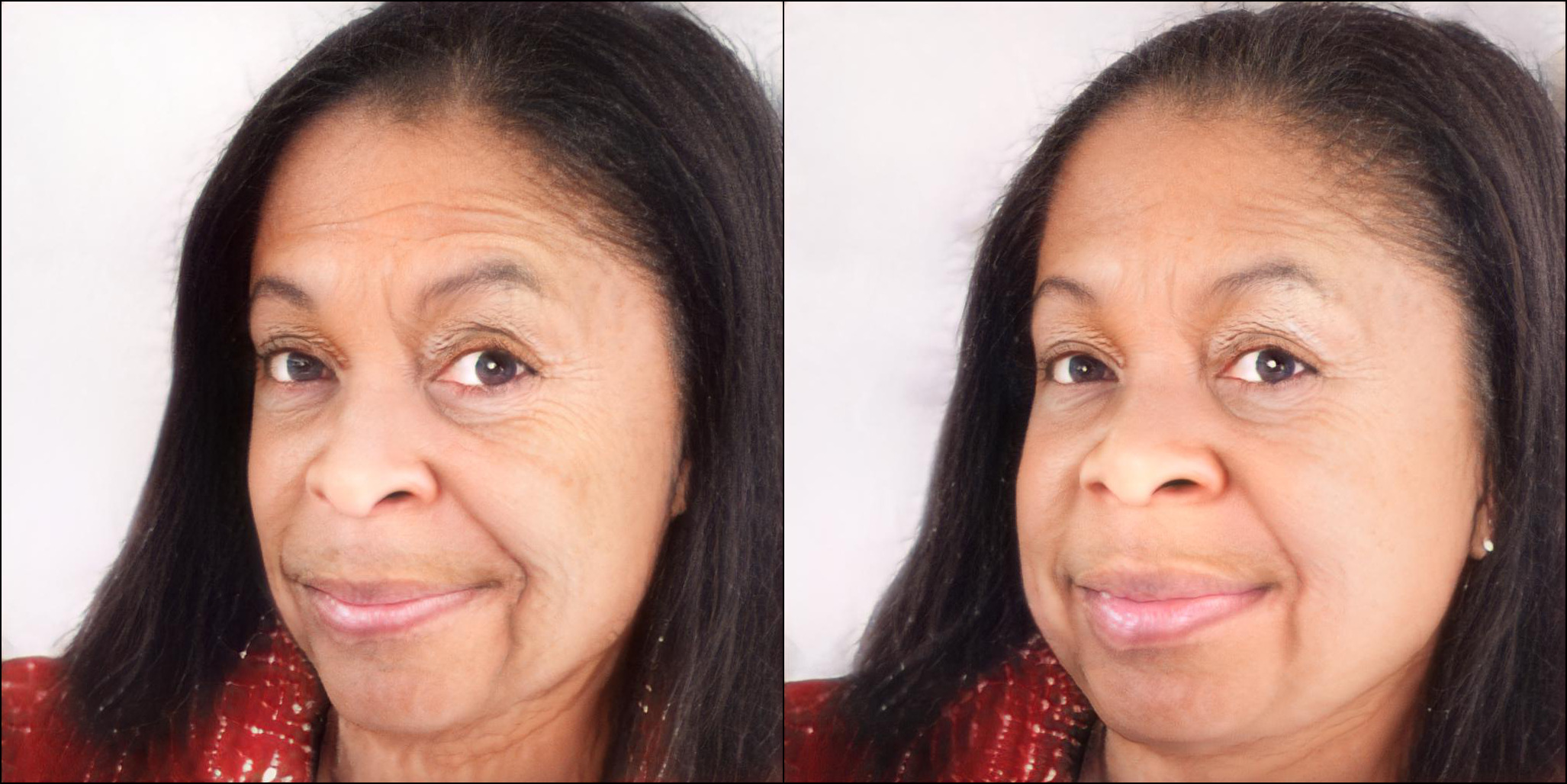} &
    \includegraphics[width=4cm]{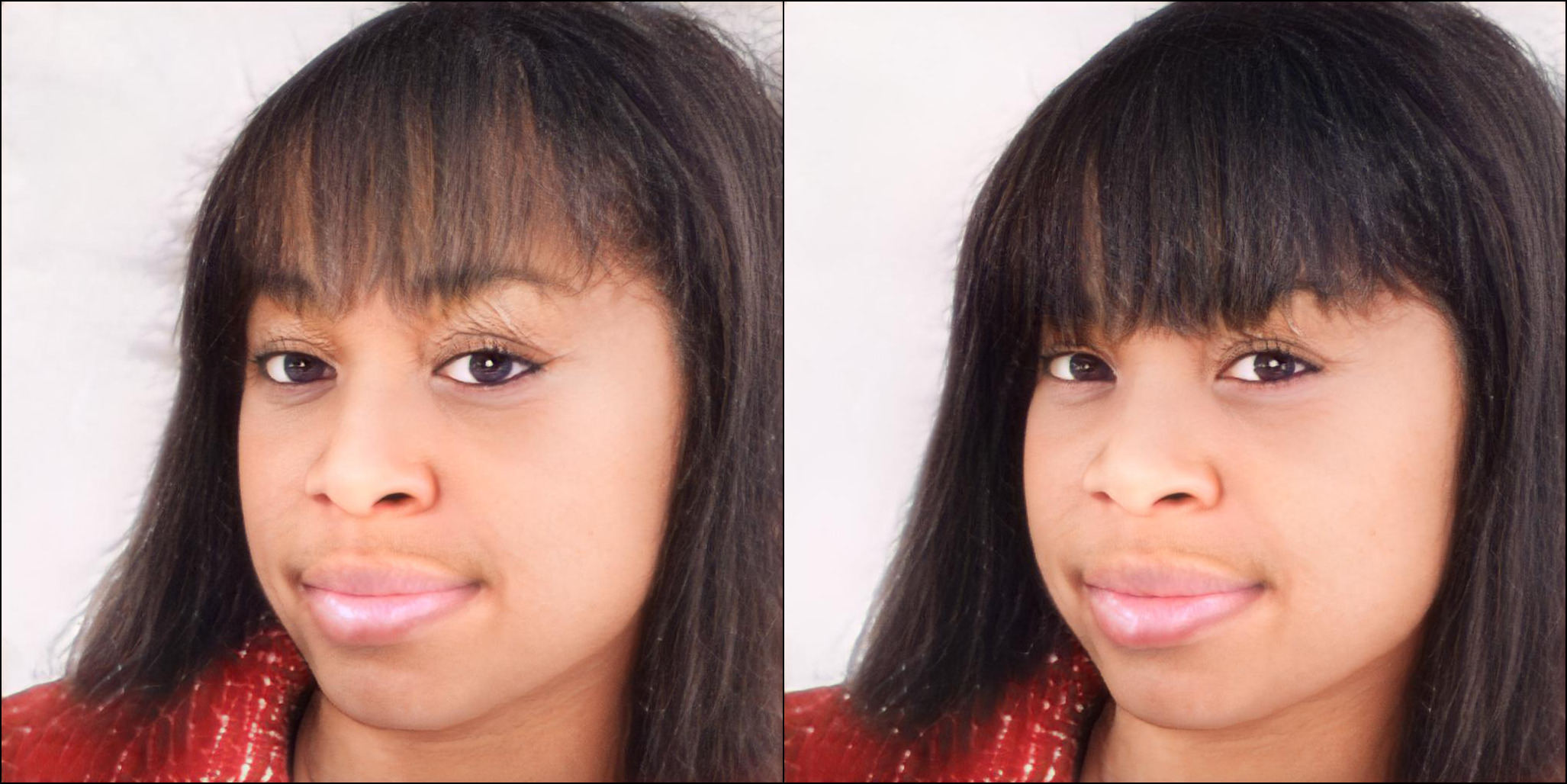} 
    & 
    \includegraphics[width=4cm]{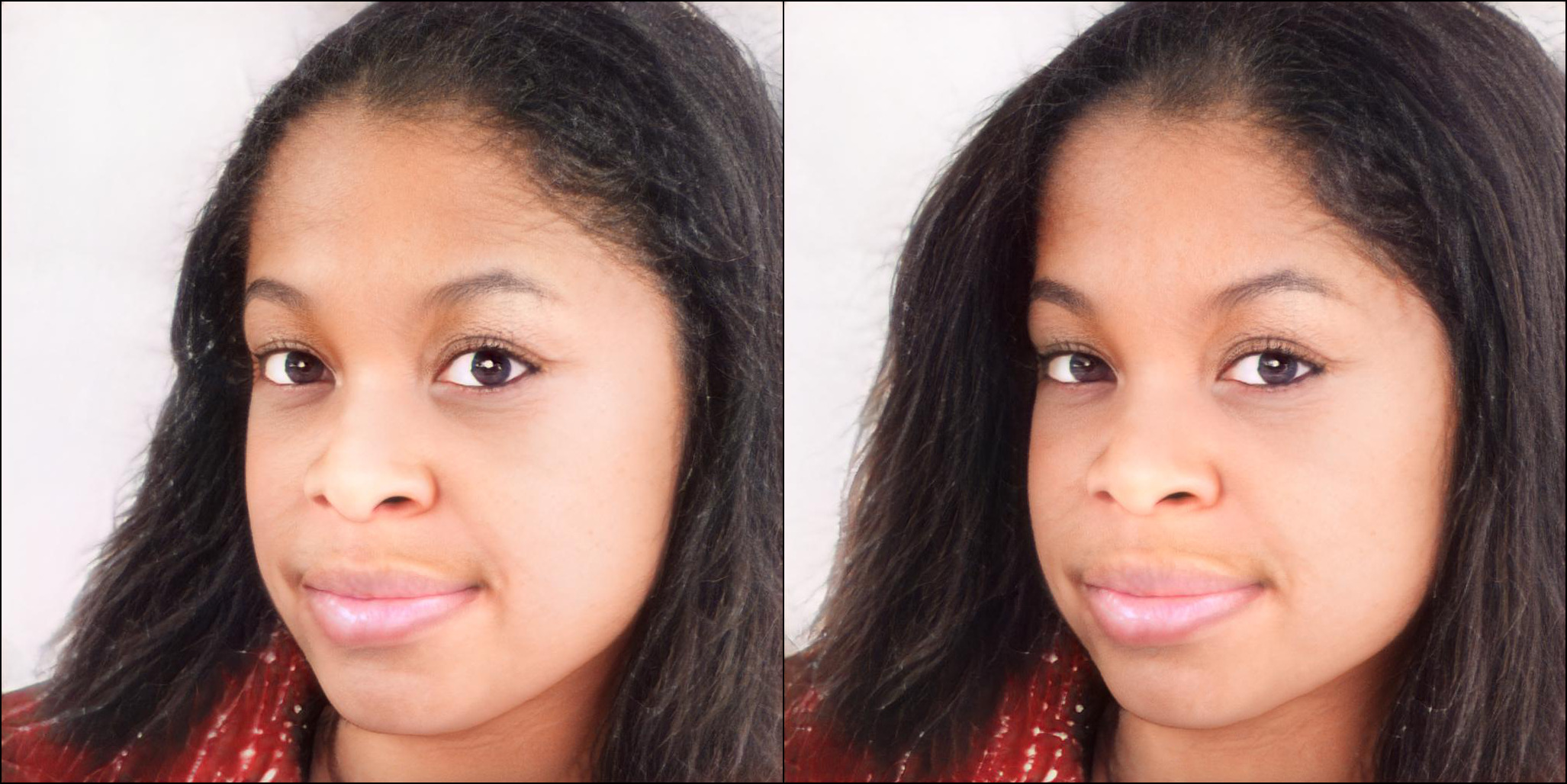} \\
    Input & LT \cite{yao2021latent} \hspace{1cm} LW (Ours) & LT \cite{yao2021latent} \hspace{1cm} LW (Ours) & LT \cite{yao2021latent} \hspace{1cm} LW (Ours) \\
    \end{tabular}
    \caption{
    \textbf{Qualitative results for facial attribute editing}. We report the editing results for $\alpha=\pm2$. We observe that our approach better preserves identity and some facial attributes (\eg expression, absence of makeup) compared to LT.
    }
    \label{fig:qualitative_results}
\end{figure*}

\begin{figure*}[t]
    \centering  
    \includegraphics[width=17cm]{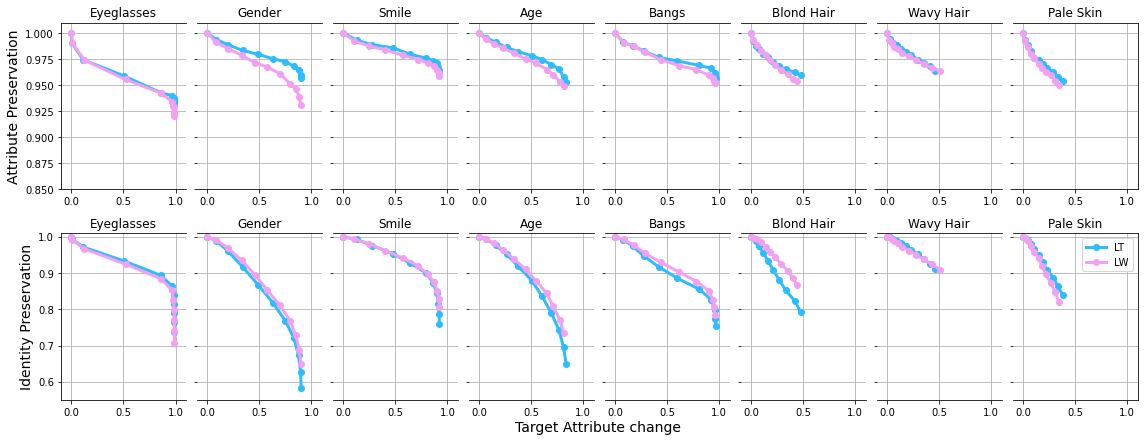}
    \caption{
    \textbf{Quantitative results for facial attribute editing}.
    We report the attribute preservation rate (computed on all other attributes indicated here) and the identity preservation rate for different values of $\alpha$ (points of the curves). The \texttt{x}-axis is the ratio of images (among all test images) for which the target attribute is successfully flipped.
    }
    \label{fig:quantitative_results_ffhq}
\end{figure*}

\subsection{Implementation Details}
We present two editing applications: facial attributes on FFHQ/CelebAHQ and number of digits on MultiMNIST\cite{mulitdigitmnist}, consisting of images with 1 to 4 MNIST digits.
We apply the editing in the latent space of StyleGAN2 \cite{Karras2019stylegan2} pretrained on FFHQ resp. MultiMNIST.
For the training data, we employ latent codes corresponding to real images previously projected in latent space using the pSp encoder \cite{richardson2021encoding}, that projects the images into the $\mathcal{W}+$ latent space. 
We employ respectively the $30K$ labeled $1024\times1024$ CelebAHQ images \cite{karras2017progressive} for face editing and $25K$ $128\times 128$ MultiMNIST images.
To learn a transformation, we use the implementation of the Wasserstein loss provided by the GeomLoss~\cite{feydy2019interpolating} library. We set the batch size as the minimum between the number of samples in the source and target distributions,  
and drop the last batch if it causes a strong imbalance between both. We use Adam optimizer with a learning rate of 0.001. To avoid overfitting the target distribution, we perform early stopping on a hold-out validation set.
As CelebAHQ contains various biases, we weight the samples and use the 
disentanglement loss. Optimal value for $\lambda$ is 1 for all considered attributes except for ``Glasses'' ($\lambda=15$). 
The cost is computed on all 40 attributes of CelebA \cite{liu2015faceattributes}. Samples are weighted based on the most common attributes.

\subsection{Metrics} 
We use three metrics \cite{yao2021latent} to evaluate the different methods.
The \emph{target attribute change} rate indicates the percentage of images for which the target attribute is indeed modified.
The \emph{attribute preservation} rate corresponds to the average number of attributes, apart from the target attribute, that are preserved. 
The aforementioned metrics are computed by running pretrained attribute image predictors before and after the editing (for a given $\alpha$) and finding which attributes have changed. An attribute is considered present if the probability is greater than 0.5. 
We also compute the \emph{identity preservation} rate as the average of the cosine similarities between ArcFace \cite{deng2019arcface} features of input 
and edited images. 
All metrics are evaluated on $1,000$ images from FFHQ. The attribute and identity preservation rates are reported against the target change for 10 values of $\alpha \in [1\cdot d, 2 \cdot d]$ where $d$ is chosen such that the target change for a given $\alpha$ is comparable between the different methods.
In tables, we report the mean over all values of $\alpha$.

\subsection{Facial Attribute Editing}

We present a quantitative and qualitative comparison with Latent Transformer (LT) \cite{yao2021latent} that relies on the guidance of a latent classifier. In addition to the classification objective, the authors introduce a disentanglement loss and an $L2$-regularization on the norm of the transformation. The latter is used to enforce identity preservation but is also critical to obtain latent codes corresponding to realistic images. The comparison is conducted on common attributes (``Glasses'', ``Gender'', 
``Smile'', ``Age'') and rarer attributes chosen based on their representation and the performances of the image classifiers (``Pale Skin'', ``Bangs'', ``Blond Hair'', ``Wavy Hair''). Quantitative results from \cref{fig:quantitative_results_ffhq} show that our results are on par with LT with occasionally slightly lower attribute preservation (``Gender'') but generally higher identity preservation (``Gender'', ``Age'', ``Blond Hair''). Note that this is surprising since we do not explicitly enforce identity preservation.
Qualitative results in \cref{fig:qualitative_results} showcase some advantages of our method.
Nose, lips and eyes shape are much better preserved for ``Gender'' and ``Age''. LT also produces ``cartoonish'' edits for these attributes while ours remains naturalistic. LT 'Gender' editing is also heavily entangled with 'Makeup' while LW adds nearly none. We provide additional qualitative results in the supplementary.

\begin{table}[t]
    \centering
    \caption{
    Quantitative results for the attributes ``Gender'' (G), ``Age'' (A) and ``Pale Skin'' (PS). We compare the classifier loss approach (LT)
    with our Wasserstein loss approach (LW). 
    Setting  ${(\text{*})}$ is the ``core'' method, w/o any regularization.}
    \begin{tabularx}{0.45\textwidth}{l | Y Y Y | Y Y Y}
    Method & \multicolumn{3}{c|}{Attr. preservation} & \multicolumn{3}{c}{Id. preservation}  \\
    \midrule
    & G & A & PS & G & A & PS \\
    LT$^{(\text{*})}$ & 0.92 & 0.95 & 0.86 & \textbf{0.94} & 0.96 & 0.96 \\
    LW$^{(\text{*})}$ &\textbf{0.95} & \textbf{0.96} & \textbf{0.96} & \textbf{0.94} & \textbf{0.97}& \textbf{0.98} \\

    \midrule
    LT  & \textbf{\ul{0.98}} & \textbf{\ul{0.98}} & \textbf{\ul{0.98}} & 0.95 & 0.96 & \textbf{\ul{0.98}} \\
    LW  & 0.97 & \textbf{\ul{0.98}} & 0.97 & \textbf{\ul{0.96}} & \textbf{\ul{0.97}} & \textbf{\ul{0.98}} \\
    \bottomrule
    \end{tabularx}
    \label{tab:quantitative_results_ablation}
\end{table}

\begin{figure}[ht]
    \centering
    \hspace{1cm} Input \hspace{1.2cm} LT$^{(\text{*})}$ 
    \hspace{1cm} LW$^{(\text{*})}$ \hspace{0.5cm} LW (Ours) \\
    \raisebox{0.3in}{\rotatebox[origin=c]{90}{\small Gender}}
    \includegraphics[width=8cm]{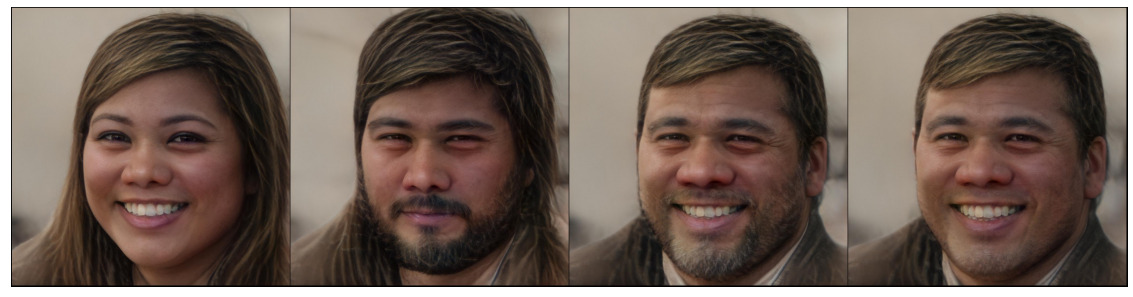} \\
    \raisebox{0.4in}{\rotatebox[origin=c]{90}{\small Age}}
    \includegraphics[width=8cm]{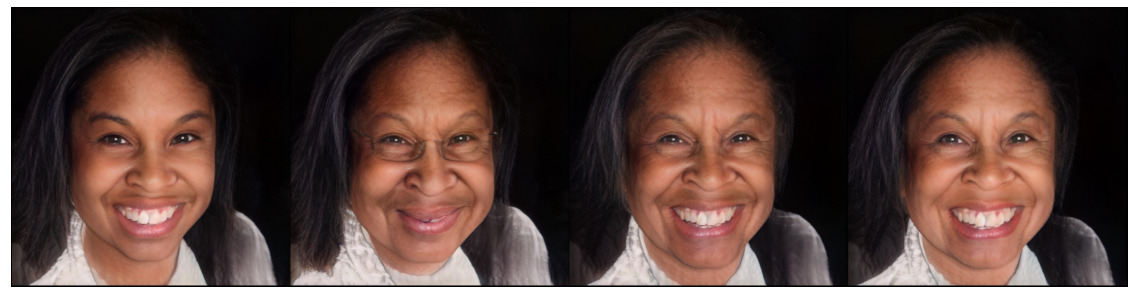} \\
    \raisebox{0.4in}{\rotatebox[origin=c]{90}{\small Pale Skin}}
    \includegraphics[width=8cm]{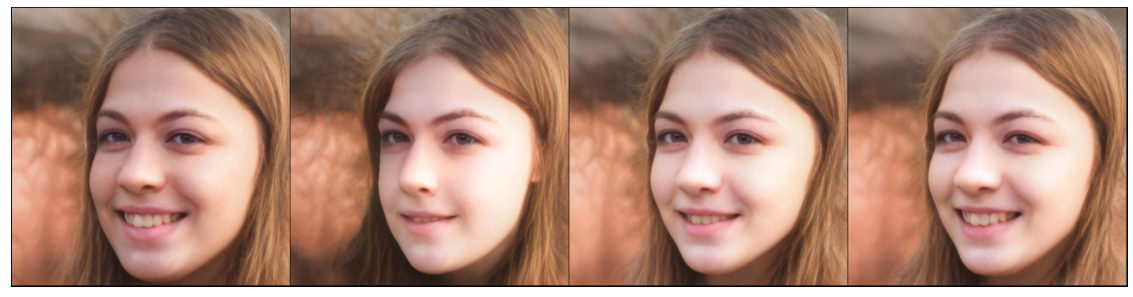}
    \caption{Qualitative comparison between classifier-based edits \big(LT$^{(\text{*})}$\big) and our Wasserstein-based approach without any constraint \big(LW$^{(\text{*})}$\big) vs with the disentanglement constraint (LW).}
    \label{fig:qual_results_ablation}
    \vspace{-0.5em}
\end{figure}

\textbf{Classifier vs.\ Wasserstein.}
We evaluate the ability of both methods to achieve disentangled and identity preserving editing without any explicit constraint. 
We denote by LT$^{(\text{*})}$ the latent transformer of Yao \textit{et al}. \cite{yao2021latent} trained without the disentanglement loss
nor the $L2$-regularization. 
In \cref{tab:quantitative_results_ablation} (top), we compare it to our model trained without the disentanglement loss ($\lambda=0$), denoted by LW$^{(\text{*})}$. The  Wasserstein baseline outperforms the classifier baseline both regarding disentanglement and identity preservation. As shown in the qualitative results presented in \cref{fig:qual_results_ablation}, the latter produces highly entangled edits (\emph{e.g.}\ with the attribute ``Smile'') and alters the identity. Without enforcing it 
explicitly, the Wasserstein 
approach already exhibits a good disentanglement ability and the identity is also well-preserved. These abilities can be explained as the Euclidean cost in employed in 
\cref{eq:l_edit} fairly reflects the perceptual distance in image space.

\textbf{Disentanglement Constraint.}
We study the influence of adding the 
disentanglement constraint from~\cref{eq:attribute_cost}. As shown in \cref{tab:quantitative_results_ablation}, we improve attribute preservation.
Qualitatively, the results are also improved as shown in \cref{fig:qual_results_ablation}. 
``Gender'' is no longer heavily entangled with ``Beard'' (1st row) and the slight entanglement with ``Smile'' is removed. 
As shown in \cref{fig:limitation_lc} (left), when the disentanglement 
constraint is used in the classifier-based approach, the edited images are unrealistic. The attribute and identity preservation curves show atypical behavior as image classifiers are disrupted by such images. As the decision boundaries of classifiers cover areas that are larger than the area of training samples, latent codes which are far away from the training distribution can still minimize the classification objective. The $L2$-regularization 
in \cite{yao2021latent} enforcing 
that the edited latent codes remain close to the initial ones is thus necessary to circumvent this limitation. Our method does not require any regularization to produce realistic edits, since our main objective enforces closeness to the target distribution.

\subsection{Editing the Number of Objects}

\begin{figure}
    \centering
    \hspace{1cm} Input \hspace{0.5cm }LT \hspace{0.5cm} LW (ours) \\ 
    \raisebox{0.3in}{1$\rightarrow 2$}\includegraphics[width=4cm]{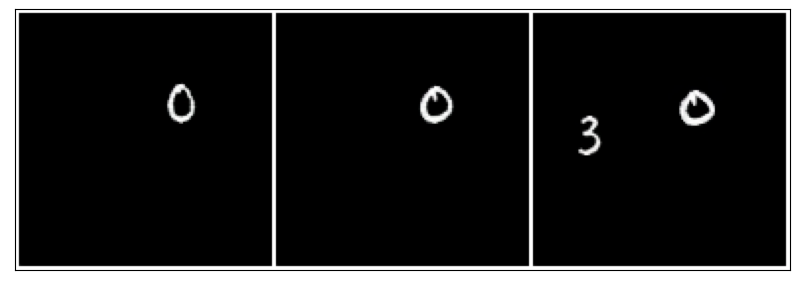} \\
    \raisebox{0.3in}{2$\rightarrow 3$}
    \includegraphics[width=4cm]{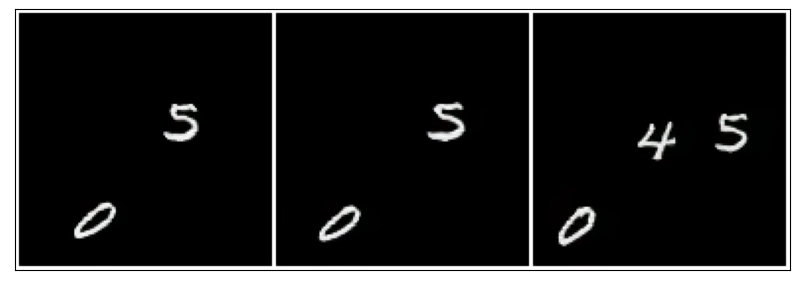} \\
    \raisebox{0.3in}{3$\rightarrow 4$}
    \includegraphics[width=4cm]{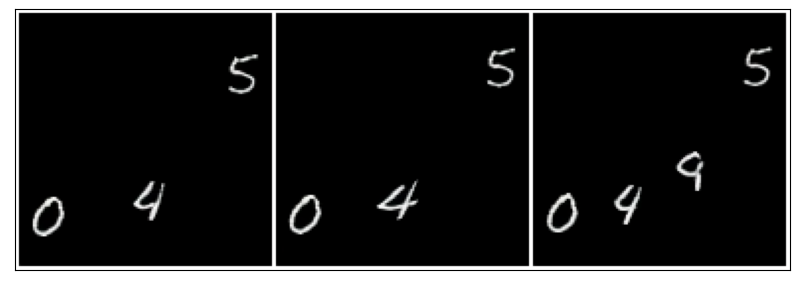}
    \caption{Qualitative results for number of objects editing. Our method adds a digit while LT fails to add one. 
    }
    \label{fig:qual_results_multimnist}
\end{figure}

The $L2$-regularization in conjunction with the classification objective is similar to the formulation employed to produce adversarial examples \cite{szegedy2013intriguing}.
While this rarely occurs on faces, this plagues editing performance on MultiMNIST.
The quantitative results in \cref{tab:multimnist_quant_results} show that for a target change of 100\% according to the latent classifier, the image classifier indicates significantly lower target changes for LT.
In other words, the latent classifier predicts that the number of digits has increased while it has stayed the same in the image, undermining the goal of an editing method.
In contrast, our method has a high editing effect and actually adds digits in the edited images. Qualitative results are provided in \cref{fig:qual_results_multimnist}.

\begin{table}[h]
\centering
\caption{Quantitative results for the manipulations ``adding one digit in an image containing $n$ digits, for $n=1,2,3$'' in real images from MultiMNIST \cite{mulitdigitmnist}. Given a target change rate of 100\% according to a latent classifier, we report the \emph{actual} change rate as measured by an image classifier. Higher values indicate a lower rate of adversarial samples.}
\begin{subtable}{0.40\textwidth}
    \begin{tabularx}{\textwidth}{c | Y Y Y}
        Method & \multicolumn{3}{c}{Target change rate} \\
        \toprule
         & 1$\rightarrow$2 & 2$\rightarrow$3 & 3$\rightarrow$ 4 \\
        LT 
        & 0.32 & 0.31 & 0.64 \\
        LW (ours) & \textbf{0.90} & \textbf{0.95} & \textbf{0.99} \\
        \bottomrule
    \end{tabularx}
\end{subtable}
\label{tab:multimnist_quant_results}
\end{table}

\section{Conclusion}

We present a new method to learn semantic editing in the latent space of GANs, that proposes to model the problem as an optimal transport problem. We look for transformations that transport a collection of latent codes to the most semantically similar points in the distribution of latent codes with the desired semantic. We use the squared Euclidean distance in latent space as a cost function as it fairly reflects the perceptual distances in image space. 
This formulation readily produces almost totally disentangled editing whereas classifier-based methods require an explicit disentanglement constraint.
To achieve even more disentangled editing, we introduce an explicit loss enforcing the transported codes to remain close to the distribution of initial codes.  
This loss is also formulated with optimal transport but using a semantic cost computed in \emph{attribute} space.
On the task of facial attribute editing on CelebA/FFHQ, our method is competitive with a state-of-the-art classifier-based method without requiring an additional constraint to ensure that the obtained images are realistic. 
Our method also alleviates other issues from using classifiers, such as the sensitivity to adversarial examples as we illustrate on the editing of the number of digits in MultiMNIST images.

Our method achieves particularly strong identity preservation performances when editing facial attributes. This is unexpected as there is no explicit constraint to do so, and the train and target distributions contain different identities. We attribute this ability to a combination of early stopping, that prevents us from overfitting our edited codes on the target distribution, and of the inductive bias of the model, which defines edits as simple affine transformations in the latent space, acting as a regularization.

While the Wasserstein loss based on the latent Euclidean distance results in state-of-the-art editing performances, it does not perfectly reflect the perceptual distance in image space. This could explain why some edits are not totally disentangled. As an extension  of this work, we believe that performances could be further improved by using a cost based  on the perceptual LPIPS metric \cite{zhang2018perceptual} or an equivalent proxy learned in the latent space to reduce computation time.

\bibliographystyle{IEEEbib}
\bibliography{references}

\end{document}